\documentclass[conference]{IEEEtran}
\IEEEoverridecommandlockouts
\usepackage{cite}
\usepackage{amsmath,amssymb,amsfonts}
\usepackage{algorithmic}
\usepackage{graphicx}
\usepackage{multirow}
\usepackage{textcomp}
\usepackage{booktabs}
\usepackage{xcolor}
\usepackage{changepage}
\usepackage[left=0.75in, right=0.75in, top=1in, bottom=0.75in]{geometry}
\usepackage{afterpage}
\usepackage{arydshln}
\def\BibTeX{{\rm B\kern-.05em{\sc i\kern-.025em b}\kern-.08em
    T\kern-.1667em\lower.7ex\hbox{E}\kern-.125emX}}

\begin{document}

\title{Dog Identification using Soft Biometrics and Neural Networks}

\author{\IEEEauthorblockN{Kenneth Lai\textsuperscript{*}, Xinyuan Tu\textsuperscript{\dag} and Svetlana Yanushkevich\textsuperscript{*}}
\IEEEauthorblockA{\textsuperscript{*}Biometric Technologies Laboratory, Department of ECE, University of Calgary, Canada \\
Email: \{kelai, syanshk\}@ucalgary.ca\\
\textsuperscript{\dag}School of Automation, Beijing Institute of Technology, China\\
Email: tuxinyuanbit@sina.com}
}
\IEEEpubid{\begin{minipage}{\textwidth}\ \\[10pt]
		\footnotesize{{\fontfamily{ptm}\selectfont Digital Object Identifier 10.1109/IJCNN.2019.8851971 \\978-1-7281-2009-6/19/\$31.00 \copyright 2019 IEEE}}
\end{minipage}}
\maketitle

\begin{abstract}
This paper addresses the problem of biometric identification of animals, specifically dogs. We apply advanced machine learning models such as deep neural network on the photographs  of pets in order to determine the pet identity.  In this paper, we explore the possibility of using different types of ``soft'' biometrics, such as breed, height, or gender, in fusion with ``hard'' biometrics such as photographs of the pet's face. We apply the principle of transfer learning on different Convolutional Neural Networks, in order to create a network designed specifically for breed classification.  The proposed network is able to achieve an accuracy of 90.80\% and 91.29\% when differentiating between the two dog breeds, for two different datasets.  Without the use of ``soft'' biometrics, the identification rate of dogs is 78.09\% but by using a decision network to incorporate ``soft'' biometrics, the identification rate can achieve an accuracy of 84.94\%.
\end{abstract}

\section{Introduction}
In the US, about 105 million people own one or more dogs, dog ownership has risen by 29\% in the past decade \cite{key}. In Canada, nearly 41\% of households have at least one dog \cite{dognumberca}. A study conducted by Weiss et al. \cite{weiss2012frequency}, found out that 14\% of dogs were lost in the past five years in the US while 7\% of them never returned to their owners. The traditional means to identify and locate lost dogs are ID collar, microchip, tattooing and GPS tag \cite{kumar2014biometric}. The first three methods include the information of the dog's name and owner's phone number which will be helpful when shelters find the lost dogs. GPS tags enable owners to locate their lost dogs directly. Nevertheless, the semi-permanent methods such as GPS tag and ID collar are vulnerable to be lost or damaged. Meanwhile, the permanent means such as microchip and tattooing are not prevalent because of their expensive prices \cite{kumar2014biometric}. In rural or remote areas of the Northern part of Canada, it is less accustomed to tattoo or tag dogs, therefore making it difficult to use the identity of the dog to claim ownership. For veterinarians, it presents a problem of  retrieving the dog's medical records and, accordingly, prevents from providing adequate healthcare.  The concept of e-health for animals is one of the major driving forces for promoting the creation of electronic medical records for pets.  For each record, a photograph of the pet is taken and can be used to verify/identify the identity of each individual pet using image processing techniques.  For example, a person who has found a dog in a remote area can take its photo using a  multimedia device and send it to a regional veterinary database for identification.

Photos of a dog head or face can be used to identify that dog, which is similar to identifying a person using the person's faces. This is a fine-grained classification which refers to classifying objects sharing similar visual features and belonging to the same basic-level class \cite{yao2012codebook}. Dog owners or veterinarians are more likely to take various pictures of dogs and store these image data for further use such as assist in finding lost dogs. There are also plenty of Apps available in the Google Play Store to find lost dogs. For instance, ``PetsFinder'' only requires a report including the location and a picture of the lost or found pet; the report is uploaded to the cloud database where owners can search for their lost pets among all the pet listings in the vicinity.

Multimedia image processing and recognition can be applied to not only identify the dog face but also breed, height and other soft biometric attributes. In this study, we investigate this approach, specifically, how classifying the breed help improve dog identification by appearance (face).
We apply the contemporary  machine learning such as deep neural networks, as well as the technique called transfer learning  \cite{torrey2010transfer} which uses a developed for one task (breed identification) to deal with another task (dog face identification).

Our major contributions are as follows:
\begin{itemize}
	\item We propose to use a coarse-to-fine process for dog recognition: we first determine the dog's breed (Section \ref{breed}) and then perform dog identification within the breed (Section \ref{dogI}).
	\item We apply transfer learning on Convolutional Neural Networks (CNN) for both the ``coarse'' and ``fine'' stages.
	\item We present several improvements to various stages: (a) at the pre-processing stage, we segment the images using their labeled facial landmarks to detect the dog's face (Section \ref{lbl:processing}); (b) at the face detection stage, we apply a Faster Region-based Convolutional Neural Networks (RCNN) dog face detector on the preprocessed images;
	(c) at the dog identification stage, we apply a combination of different kinds of CNN architectures in order to compare the identification accuracy (Section \ref{dogI}).
\end{itemize}
			 \newgeometry{left=0.75in, right=0.75in, top=0.75in, bottom=0.75in}
\begin{itemize}
\item We suggest using ``soft'' biometrics in order to  improve the performance of the classifiers (Section \ref{bn}).
\end{itemize}

\section{Related Work}
In the area of dog biometrics, ongoing research heavily focuses on using pattern recognition for dog identification as well as breed classification.

To examine the performance of recognizing different dog breeds using assorted images, Khosla et al. created a dataset of different dog breeds called Stanford Dogs Dataset \cite{khosla2011novel}.  Using Stanford Dogs Dataset, Sermanet et al. proposed the use of an attention-based model for breed classification which achieved an accuracy of 76.80\% \cite{sermanet2014attention}.  Another model using depth-wise separable convolutions was proposed in \cite{howard2017mobilenets} and yielded 83.30\% accuracy.

Similarly, another dataset, Columbia Dogs Dataset, containing images of different dog breeds was created by Liu et al. \cite{liu2012dog}. In \cite{liu2012dog}, a breed classification rate of 67.00\% can be achieved by using a combination of grey-scale SIFT descriptors and color histogram features to train an SVM. Furthermore, \cite{wang2014dog} uses the labeled landmark metadata to improve the accuracy to 96.50\% by incorporating Grassmann manifold to help distinguish the different dog breeds using their face geometries.

Pet breed classification using both shape and texture  was proposed in \cite{parkhi2012cats}. A deformable part model was used in  \cite{parkhi2012cats} to detect the shape and a bag-of-words model to capture the appearance. For a dataset that includes 37 breeds of cats and dogs,  an accuracy of 69.00\% was reported.

Paper \cite{kumar2016monitoring} was one of the first to describe dog face identification. It used Fisher Linear Projection and Preservation with One-Shot Similarity for matching and reported 96.87\%  rank-4 accuracy. The reported dataset is not publicly available, thus making it difficult to perform a proper comparison.

Another research \cite{chen2016locality} was conducted in 2016 aiming at cat face identification by exploiting visual information of cat noses. They designed a representative dictionary with data locality constraint based on a dataset containing 700 cat nose images from 70 cats. This method reached an accuracy of 91.20\%.

Moreira et al. evaluate the viability of using existing human face recognition methods (EigenFaces, FisherFaces, LBPH, and a Sparse method) as well as deep learning techniques such as Convolutional Neural Networks (BARK and WOOF) for dog recognition \cite{moreira2017my}.  Using a dataset of two different breeds of dogs, huskies and pugs, Moreira et al. show that based on using the WOOF model, an accuracy of 75.14\% and 54.38\% is obtained for huskies and pugs, respectively.

\section{Methodology and approach}
\begin{figure*}
\begin{center}
  \includegraphics[width=0.99\linewidth]{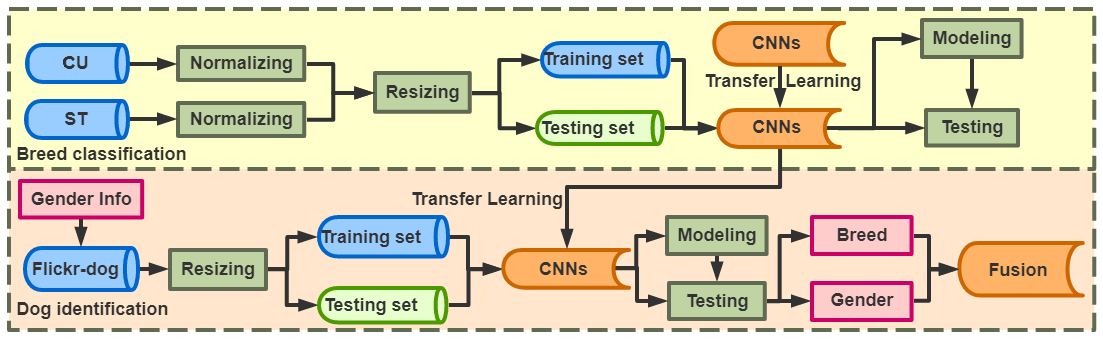}
\end{center}
  \caption{Structure of the proposed framework. The upper part implements a ``coarse'' stage which is dog breed classification given a probe image of a dog; the lower part is the ``fine'' stage to match the probe image within the subset of the detected dog breed or top $k$ similar breeds.}
\label{figure:pipeline}
\end{figure*}

In this paper, we propose a framework that performs both dog identification and dog breed classification.  In Figure \ref{figure:pipeline}, we illustrate the two different structures that can be used independently for identification and breed classification as well as the combined network resulting in a coarse-to-fine approach for improving the identification rate.  The top portion of the framework is designed for breed classification and the performance is examined using Stanford Dog and Columbia Dog Datasets while the bottom portion of the framework is created for dog identification and is examined using the Flickr-dog dataset.

Evaluation of the system is based on three different datasets, Columbia Dogs Dataset (CU), Stanford Dogs Dataset (ST), and Flickr-dog dataset.  Depending on the portion of the network chosen, the output of the system is either the predicted identity of the dog or the estimated breed of the dog.  In this paper, we propose to use the coarse-to-fine approach to improve the accuracy of predicting the identity of the dog based on using ``soft'' biometrics such as the dog's breed.

\subsection{Dataset}
\begin{figure}[!htb]
\begin{center}
  \includegraphics[width=0.95\linewidth]{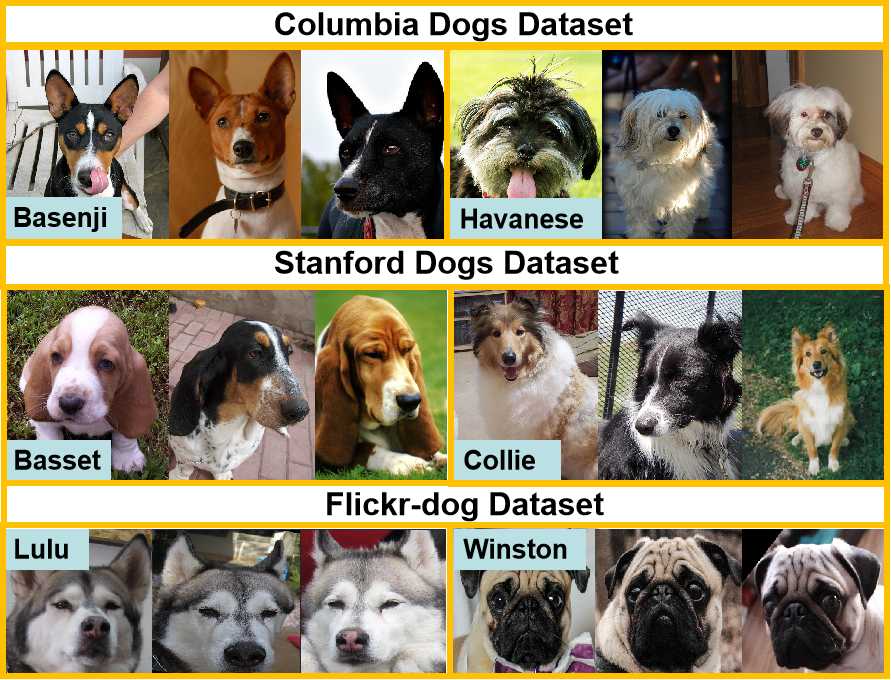}
\end{center}
  \caption{Images from the three databases: CU and ST of various dog breeds, and  Flickr-dog of  individual dogs of two breeds (pugs and huskies)}
												\label{figure:database}
\end{figure}

We employed three datasets in our work: CU, ST, and Flickr-dog dataset. CU and ST were used for breeds classification while Flickr-dog was considered for dog identification. Sample images in these three datasets are shown in Figure \ref{figure:database}.
\subsubsection{Stanford Dogs Dataset} \label{STdogs}

Stanford Dogs Dataset \cite{khosla2011novel} includes 20,580 images covering 120 breeds with 8,580 images for testing and 12,000 images for training.  The dataset is built using the images and annotation from ImageNet.  The annotation files indicate the bounding boxes for the body of the dogs in each image.  There are approximately 150 images per each breed and the images for each breed do not have a consistent image resolution.
\subsubsection{Columbia Dogs Dataset} \label{CUdogs}

Columbia Dogs Dataset \cite{liu2012dog} contains 8,351 images belonging to 133 breeds, 3,575 images for testing and 4,776 images for training. This dataset is gathered from Google image search, ImageNet, and Flickr.  For each image, the coordinates of eight key points are annotated: right eye, left eye, nose, right ear tip, right ear base, top of head, left ear base, and left ear tip. For this dataset, each image can have widely different image resolution and the images belonging to the same breed can have significantly different locations of key points, fur color, or pose.
\subsubsection{Flickr-dog Dataset} \label{Fdogs}

Flickr-dog dataset \cite{moreira2017my}  contains two breeds of dogs: pugs and huskies. The dataset contains 42 classes, 21 for each dog breed, up to a total of 374 images with at least 5 images per class. In addition, the images in the dataset are normalized such that all images are cropped, aligned, and resized.  Each image contains a horizontally aligned face of the dog with an image resolution of $250 \times 250$ pixels.
\subsection{Data processing methods and tools} \label{lbl:processing}

The framework shown in Figure \ref{figure:pipeline} represents the data processing and classification stages required in the coarse-to-fine technique:
\begin{enumerate}
	\item Stage 1: Normalizing and resizing of images.
	\item Stage 2: Splitting dataset into training and testing sets.
	\item Stage 3: Classifying breeds and recognizing identity using CNN classifiers.
\end{enumerate}
 
The ``coarse'' step, breed classification,  detects the top $k$ possible breeds for the given probe image of a dog. The parameters of the CNNs are tuned to obtain the optimized network. The ``fine'' stage represents the process of identifying the identity of the dog by comparing the probe image against the gallery of predicted top-ranked breeds. Transfer learning can be applied to transfer the parameters of the CNNs to focus on the task of finding the identity of the dog as opposed to the breed.

We normalized the images in CU and ST using the proposed method described in Section \ref{lbl:processing}. Then we used the normalized images in CU to train a Faster RCNN dog face detector.  Using the Faster RCNN, we create a new dataset comprised of cropped dog face images from both the CU and ST dataset.

The images of the top-$k$ breeds for the unidentified dog were supplied to the different CNNs whose parameters are fine tuned for dog identification. Since the databases used for breed classification were limited in the number of images, specifically very few images from the same dog, we chose to use Flickr-dog, a database that contains multiple images from the same dog. Because Flickr-dog only contains two breeds of dogs, huskies and pugs, we made an assumption that these two selected breeds were the ``top-two'' results from the ``coarse'' stage and used these results for dog identification. We compared the performance of different CNNs including, InceptionV3, MobileNet, VGG-16, and Xception.  

\begin{figure}[!htb]
\begin{center}
  \includegraphics[width=0.8\linewidth]{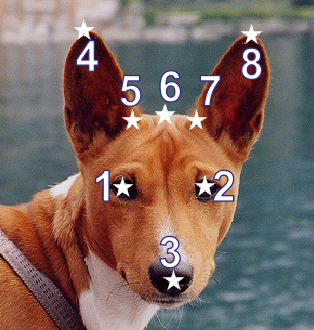}
\end{center}
  \caption{Eight points labeled on the  dog images in the Columbia Dogs Dataset}
											\label{figure:8}
\end{figure}

The normalization of images from the CU was composed of two steps: rotation and segmentation. Every picture in the database was labeled using eight key points as shown in Figure \ref{figure:8}.

The normalization step is as follows: we calculate the angle ($\theta$) between the $x$-axis and the line connecting the dog's eyes. Based on this angle, we rotate the image in order to align the eyes horizontally.

Face detection and cropping are achieved in the following steps:
\begin{itemize}
		\item We approximate the face width ratio based on the labeled points of the face. In the CU, the distance between labels 1 and 2 is 1/3 of the face width.
	\item  We define the centroid $c$ as the point in the triangle created from the two eyes and the nose. The distance between points $c$ and 6 is 1/2 of the face length, excluding the ears.
	\item We determined whether the ears ``stand up'' or ``hang down'' by comparing the $y$ coordinate value of point 6 and the smaller $y$ value among points 4 and 8. If at least one of the ears stands up, the distance between the top of the head and the standing ears is taken into consideration when calculating the total face length.
	\item The estimated length and width are used to form a rectangle enclosing the dog's face.
\end{itemize}

\begin{figure}[!htb]
\begin{center}
  \includegraphics[width=0.9\linewidth]{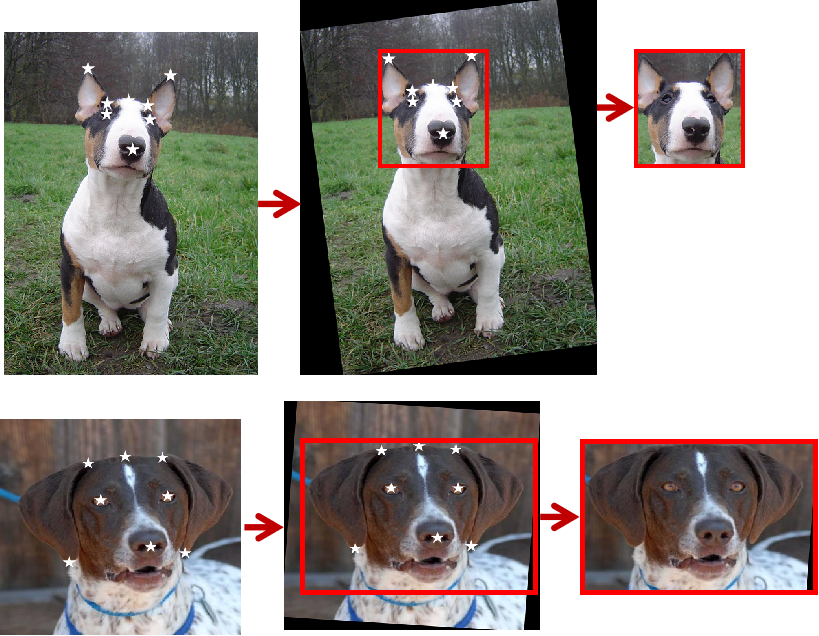}
\end{center}
  \caption{Illustration of cropping of the dog face for two dogs with various ear positions}
											\label{figure:crop}
\end{figure}

 Two examples of applying the above steps are shown in Figure \ref{figure:crop}.

For the ST, the normalization operation is based on cropping of the dog body using the provided annotated region of interest.
\subsection{Dog face detection}

\begin{figure}[!htb]
\begin{center}
  \includegraphics[width=0.95\linewidth]{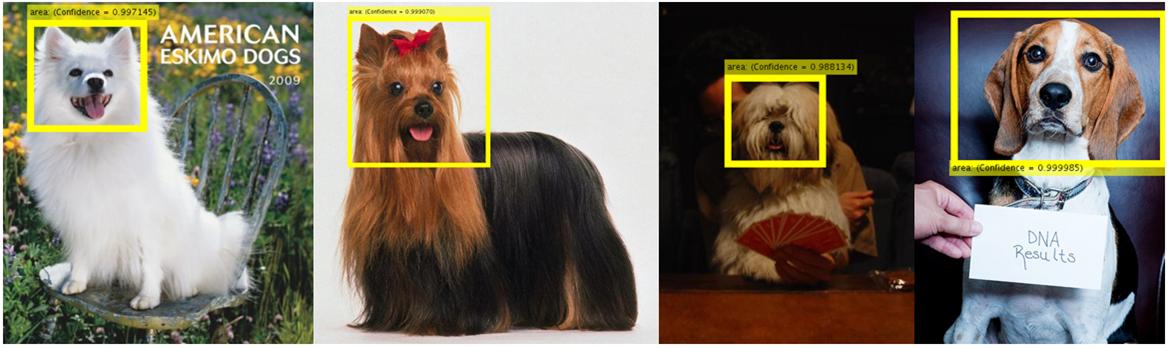}
\end{center}
  \caption{Results of the Faster RCNN Detector on dog images with various ear positions}
											\label{figure:rcnn}
\end{figure}

We trained a Faster RCNN \cite{ren2015faster} Face Detector based on the VGG-16 \cite{simonyan2014very} architecture with the images of dog faces based on our proposed face cropping method. Then we applied this detector on CU and ST to obtain only faces images of dogs. Some results of the face detector are shown in Figure \ref{figure:rcnn}.

\subsection{``Coarse stage'': Breed Classification}

After preprocessing, each database is divided into a training set and a testing set. The training set images are resized to $224 \times 224$ or $299 \times 299$ pixels depending on the selected CNN.  In this paper, we choose to examine InceptionV3 \cite{szegedy2016rethinking}, MobileNet \cite{howard2017mobilenets}, VGG-16 \cite{simonyan2014very}, and Xception \cite{chollet2017xception} because these networks all have the ability to classify 1000 object categories based on the large scale of training done for the ImageNet challenge. Using transfer learning, we incorporated these networks with their trained weights with exception of the top-classification layers which were replaced with an average pooling layer, two fully connected layers, and a ``softmax'' layer for classification. We also adjusted the output of the fully connected layer to be the number of breeds. These networks have learned rich feature representations for a wide range of images which equips it with strong generalization ability to extract features from images. After training the CNNs with the selected training sets, the weight becomes more fine-tuned for extracting features from images of dogs.

\subsection{``Fine stage'': Dog Identification}

After determining the top $k$ possible breeds of the unidentified dog, we can narrow our scope to these possible breeds for dog identification.  This process can greatly improve the identification accuracy as the pool of possible identity candidates is greatly reduced based on the breed. For this part, only Flickr-dog dataset can be used because it is the only dataset with labeled information on dog identities. For identification, multiple images from the same dog identity are necessary in order to create a gallery of training and probing images.  As the Flickr-dog dataset only contains identities for two breeds of dogs, we designed an experiment that assumes the estimated breed classification result (from ''coarse stage'') as either husky or pug. The dog identification is then performed for each  breed separately.
\subsubsection{Data Augmentation}

Since the number of images in Flickr-dog is limited, data augmentation is required to create a larger training pool for the training phase. For data augmentation, we choose to use reflection with respect to the vertical axis as a way to double the number of images.

\subsubsection{Identification}

Similarly to breed classification, we selected different CNNs and applied transfer learning to fine-tune parameters for recognition of the identity of the dog. For each of the CNNs, we discarded the top-classifying layers which were tuned for the ImageNet and replaced these layers with new full connected layers in order to use these networks to recognize the identity of dogs.

We separated the Flickr-dog set into training sets and testing sets following the approach proposed in \cite{moreira2017my}. We performed 5-fold cross-validation to compute the accuracy of identification. We randomly split the pictures of each dog into 5 groups: 4 groups were used for training and the remaining group was used for testing. The process was repeated until every group has been tested.

\section{Experimental Results}
In our experiment, we separated CU, ST, and the combined dataset into the training  and testing sets.  We trained the CNNs on five input datasets:  raw images in CU, normalized images in CU, raw images in ST, normalized images in ST, and the images in the combined dataset. For the combined database, we used 16,776 (12,000 + 4,776, the training set from ST and CU) of the images for training and 12,155 (8,580 + 3,575, testing set from ST and CU) for validation which were separated based on the original test/training set specified in the CU and ST datasets.  We tested the performance of breed classification for the different CNNs using selected testing sets for each database.  

For dog identification, we applied a transfer learning on the CNNs. We estimated the accuracy of identification using cross-validation.

\subsection{Breed Classification} \label{breed}

First, we trained the CNNs based on two groups of training data. The first group is the raw images from the CU and its normalized images. The other group is the raw images from the ST and its normalized images. Similarly, there were two groups of testing sets to test the classifiers. In addition, we applied a decision-level fusion to the probabilities obtained from the raw and normalized images by applying an $\alpha$ weight to the probability for each decision. As there are only two decisions, the weights are applied to the confidence of each decision as opposed to the decision itself.  Since we are performing fusion between two different datasets, we choose to apply a ``weighted sum'' fusion as follows: 
\begin{equation}\label{eq:fusion}
P(Decision)=\alpha \times P(D_{raw})+(1-\alpha)\times P(D_{normalized})
\end{equation}

The performance metrics, the averaged accuracy, and the balanced accuracy are defined as follows:
	\begin{eqnarray} \label{eq:acc-avg} 
	average = \frac{TP}{FN + TP} 
	\\\label{eq:acc-bal} 
	balance= \frac{1}{N} {\sum^{N}_{i}\frac{TP_i}{FN_i + TP_i}} 
	\end{eqnarray}
where $TP$ is the number of correctly classified images using rank-one matching, $FN$ is the number of incorrectly classified images,  and $FN + TP$ is the total number of images.  For the balanced accuracy, $i$ indicates a selected class and $N$ is the total number of classes.   The balanced accuracy (Equation \ref{eq:acc-bal}) reports the averaged accuracy between each class of breed regardless of the number of images within each breed; whereas averaged accuracy (Equation \ref{eq:acc-avg}) is based on the total number of breeds predicted correctly.  Both the balanced and averaged accuracy should be identical if the number of images evaluated for each class is equivalent; however, since there is an unbalanced number of images for each class, there is a slight discrepancy between these accuracies.
The accuracy  of these experiments and other classifiers reported in the literature for the same datasets is shown in Table \ref{table:1}.

\begin{table}[!htb]
	\caption{Accuracy of dog breed classification based on three different datasets (CU, ST and ST+CU) for various network models}
\begin{center}
\begin{tabular}{c|c||rrrrr}
\parbox[t]{2mm}{\multirow{2}{*}{\rotatebox[origin=c]{90}{Data}}} & \multirow{2}{*}{Model} & \multicolumn{5}{c}{Accuracy} \\
& &	\multicolumn{1}{c}{Worst}	&	\multicolumn{1}{c}{Best}	&	\multicolumn{1}{c}{$\sigma$}	&	\multicolumn{1}{c}{Avg.}	&	\multicolumn{1}{c}{Bal.}	\\
	\hline
	\hline
\parbox[t]{2mm}{\multirow{4}{*}{\rotatebox[origin=c]{90}{CU-Norm}}}	&	InceptionV3	&	23.53	&	100.00	&	15.61	&	81.29	&	80.00	\\
&	MobileNet	&	11.11	&	100.00	&	15.94	&	76.03	&	74.86	\\
&	VGG-16	&	26.67	&	100.00	&	17.95	&	71.72	&	69.78	\\
&	Xception	&	27.78	&	100.00	&	14.34	&	\textbf{83.58}	&	\textbf{82.28}	\\
\hdashline													
\parbox[t]{2mm}{\multirow{4}{*}{\rotatebox[origin=c]{90}{CU}}}	&	InceptionV3	&	27.78	&	100.00	&	12.69	&	86.99	&	85.96	\\
&	MobileNet	&	16.67	&	100.00	&	15.59	&	80.84	&	79.22	\\
&	VGG-16	&	13.33	&	95.65	&	17.85	&	62.57	&	60.49	\\
&	Xception	&	33.33	&	100.00	&	11.82	&	\textbf{89.01}	&	\textbf{88.03}	\\
\hdashline													
\parbox[t]{2mm}{\multirow{4}{*}{\rotatebox[origin=c]{90}{Fusion}}}&	InceptionV3	&	22.22	&	100.00	&	12.46	&	89.15	&	87.99	\\
&	MobileNet	&	16.67	&	100.00	&	14.21	&	84.64	&	83.57	\\
&	VGG-16	&	17.65	&	100.00	&	17.46	&	74.01	&	71.88	\\
&	Xception	&	27.78	&	100.00	&	11.71	&	\textbf{90.80}	&	\textbf{89.88}	\\
\hdashline
	& Wang \cite{wang2014dog} & \multicolumn{1}{c}{--} & \multicolumn{1}{c}{--}& \multicolumn{1}{c}{--} & \multicolumn{2}{c}{96.50} \\
	& Liu \cite{liu2012dog} & \multicolumn{1}{c}{--} & \multicolumn{1}{c}{--}& \multicolumn{1}{c}{--} & \multicolumn{2}{c}{67.00} \\
\hline													
\parbox[t]{2mm}{\multirow{4}{*}{\rotatebox[origin=c]{90}{ST-Norm}}}	&	InceptionV3	&	52.00	&	100.00	&	9.98	&	87.47	&	86.95	\\
&	MobileNet	&	40.00	&	100.00	&	12.22	&	78.07	&	77.35	\\
&	VGG-16	&	31.25	&	94.83	&	13.73	&	64.49	&	63.85	\\
&	Xception	&	52.00	&	100.00	&	9.17	&	\textbf{90.03}	&	\textbf{89.62}	\\
\hdashline													
\parbox[t]{2mm}{\multirow{4}{*}{\rotatebox[origin=c]{90}{ST}}}	&	InceptionV3	&	48.00	&	100.00	&	9.67	&	86.76	&	86.42	\\
&	MobileNet	&	40.00	&	98.28	&	11.58	&	75.57	&	75.13	\\
&	VGG-16	&	27.78	&	87.93	&	13.15	&	60.19	&	59.37	\\
&	Xception	&	54.35	&	100.00	&	8.55	&	\textbf{88.81}	&	\textbf{88.54}	\\
\hdashline													
\parbox[t]{2mm}{\multirow{4}{*}{\rotatebox[origin=c]{90}{Fusion}}}&	InceptionV3	&	56.00	&	100.00	&	8.96	&	89.70	&	89.32	\\
&	MobileNet	&	42.00	&	100.00	&	11.65	&	80.85	&	80.24	\\
&	VGG-16	&	30.56	&	94.83	&	13.81	&	65.90	&	65.05	\\
&	Xception	&	56.52	&	100.00	&	8.30	&	\textbf{91.29}	&	\textbf{90.99}	\\
\hdashline
	& Sermanet \cite{sermanet2014attention} & \multicolumn{1}{c}{--} & \multicolumn{1}{c}{--}& \multicolumn{1}{c}{--} & \multicolumn{2}{c}{76.80} \\
	& Howard \cite{howard2017mobilenets} & \multicolumn{1}{c}{--} & \multicolumn{1}{c}{--}& \multicolumn{1}{c}{--} & \multicolumn{2}{c}{83.30} \\
	\hline
\parbox[t]{2mm}{\multirow{4}{*}{\rotatebox[origin=c]{90}{ST+CU}}}	&	InceptionV3	&	0.00	&	100.00	&	23.33	&	78.39	&	69.94	\\
&	MobileNet	&	7.41	&	95.12	&	19.34	&	70.14	&	64.05	\\
&	VGG-16	&	0.00	&	94.87	&	19.91	&	55.09	&	49.05	\\
&	Xception	&	0.00	&	100.00	&	22.23	&	\textbf{81.42}	&	\textbf{73.92}	\\
\end{tabular}
\end{center}
	\label{table:1}
\end{table}

Table \ref{table:1} reports the accuracy of breed classification for various network models applied on different datasets, as well as assorted partitions for each dataset.  The ``worst'' and the ``best'' accuracy represents the worst and the best performance obtained between the different breeds of dogs, respectively.  The standard deviation ($\sigma$) indicates the spread of accuracy between the breeds.  The average accuracy (Avg.) and balanced accuracy (Bal.) are calculated using Equation \ref{eq:acc-avg} and \ref{eq:acc-bal}. 

The reported results show that the Xception CNN performs the best regardless of the dataset used. As shown in Table \ref{table:1}, using the normalized images, the breed classification accuracy is lower than the raw images; however, a fusion between the results  for the raw and the normalized images yields an improvement.  The fusion based on Equation \ref{eq:fusion} represents a weighted average between the two decisions from each image.  This results in an improved classification  because the features extracted from normalized images are different from the raw images.  Through experimentation, we found that $\alpha=0.5$ yields the best performance. 

Wang et al. \cite{wang2014dog} reported an extremely high accuracy for the CU because they utilized the labeled 2-dimensional landmarks extracted from dog faces. However, the Flickr-dog Dataset does not include the information of key points on the face. Thus, the 2D landmark method cannot be used to improve the identification performance.

\subsection{Dog Identification} \label{dogI}

The Flickr-dog dataset was chosen for the task of recognizing the identity of dogs, as it contains 42 dogs, each having multiple images from same the subject/dog.  Data augmentation was applied to increase the amount of data, and the features were extracted using different CNNs. 

We evaluated the accuracy of the classifiers for each of the breed groups (pugs and husky), as well as the complete Flickr-dog dataset using the approach proposed in \cite{moreira2017my}. We performed 5-fold cross-validation, and the final accuracy is the arithmetic mean of the five accuracies. The results for various classifiers are shown in Table \ref{table:3}.

\begin{table}[!htb]
	\caption{Accuracy of dog identification using Flickr-dog dataset for various network models}
	\begin{center}
		\begin{tabular}{c||cccccc}
			\multirow{2}{*}{Method} &  \multicolumn{2}{c}{Husky}  &  \multicolumn{2}{c}{Pug} & \multicolumn{2}{c}{Pug and Husky} \\
			&  Avg.  &  Bal. &  Avg.  &  Bal. &  Avg.  &  Bal. \\
			\hline
			\hline
InceptionV3	&	77.70	&	76.11	&	62.06	&	56.07	&	74.26	&	71.92	\\
MobileNet	&	\textbf{89.41}	&	\textbf{88.73}	&	\textbf{74.81}	&	\textbf{70.44}	&	75.38	&	73.75	\\
VGG-16		&	77.26	&	74.05	&	65.98	&	63.93	&	64.62	&	62.62	\\
Xception	&	84.50	&	83.81	&	71.58	&	65.75	&	\textbf{78.09}	&	\textbf{76.53}	\\
\hdashline
			Moreira \cite{moreira2017my} &  \multicolumn{2}{c}{75.14} & \multicolumn{2}{c}{54.38} &\multicolumn{2}{c}{66.90} \\
		\end{tabular}
	\end{center}
	\label{table:3}
\end{table}

MobileNet delivered the best performance for the individual breeds of dogs, while Xception provided the highest accuracy when both breeds of dogs are combined into one dataset.  Each of the experimented CNNs shows an improvement over Moreira et al. \cite{moreira2017my}. Specifically, Xception shows a 10-12\% increase when no breed information is provided.  MobileNet provided an increase of approximately 14\% and 20\% for recognizing the identity of husky and pug images, respectively.

\begin{figure}[!htb]
	\begin{center}
		\includegraphics[width=0.95\linewidth]{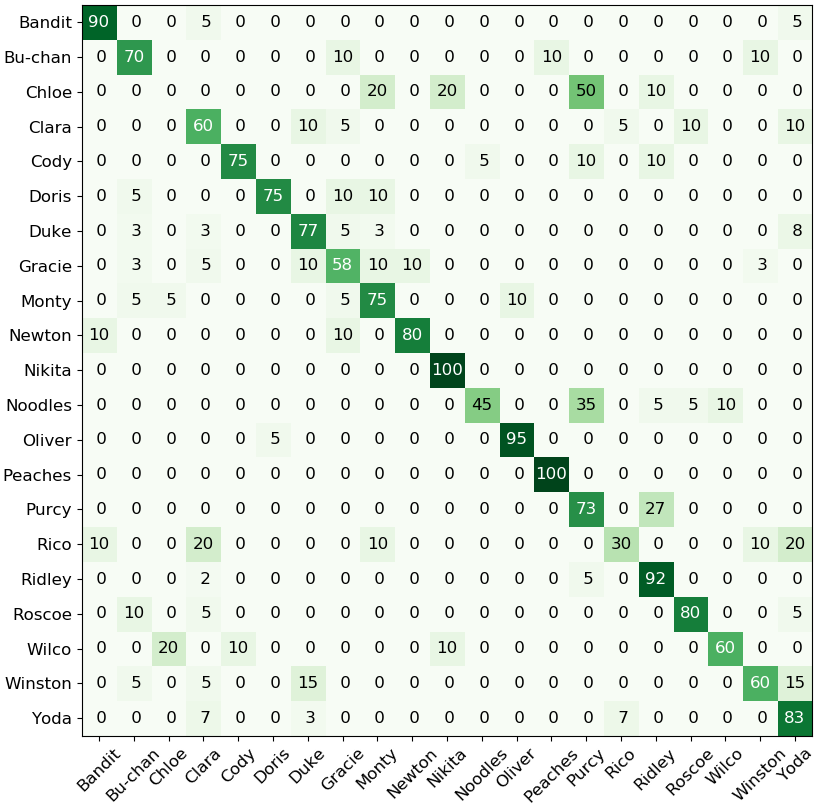}
	\end{center}
	\caption{Confusion Matrix using MobileNet for pugs only.}
	\label{figure:mobileNetPug}
\end{figure}

\begin{figure}[!htb]
	\begin{center}
		\includegraphics[width=0.95\linewidth]{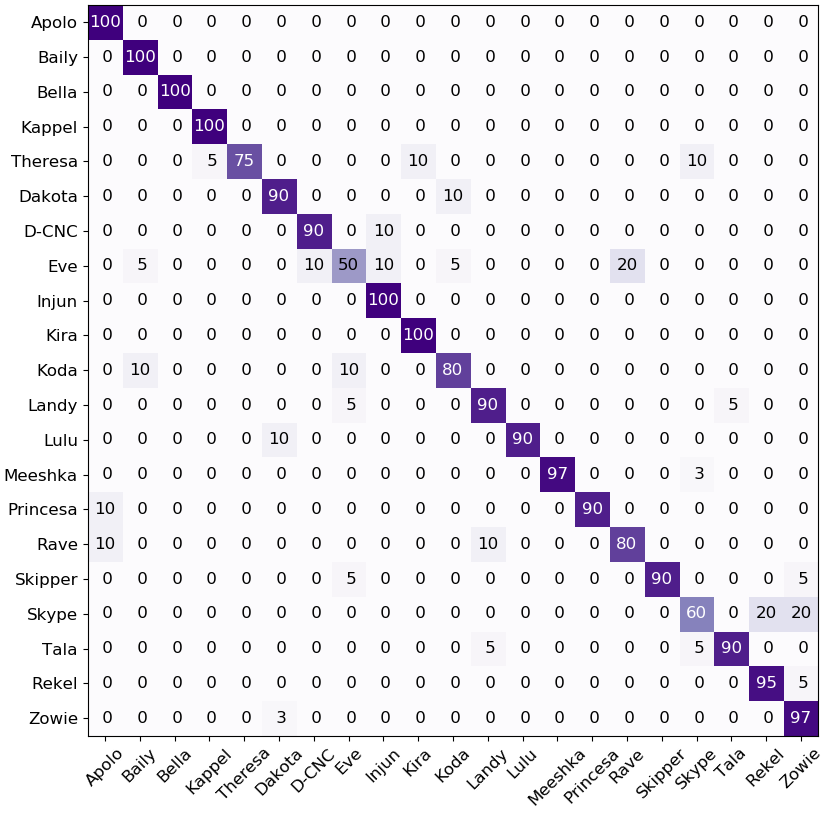}
	\end{center}
	\caption{Confusion Matrix using MobileNet for huskies only.}
	\label{figure:mobileNetHusky}
\end{figure}

To further illustrate the spread of accuracy between the identities of dogs, we generated a confusion matrix for selected huskies and pugs (Figure \ref{figure:mobileNetPug} and \ref{figure:mobileNetHusky}). Both Figure \ref{figure:mobileNetPug} and \ref{figure:mobileNetHusky} describes the performance of the dog recognition model using MobileNet.  Each row in the matrix indicates the known identity; each column shows the predicted identity.  Together, the matrix illustrates how a known identity is classified or mis-classified.  For example, in Figure \ref{figure:mobileNetHusky}, ``Eve'' is classified correctly 50\% of the time and is mis-classified as ``Rave'' 20\% of the time.

\subsection{Further Improvements}\label{bn}

\begin{figure}[!htb]
	\begin{center}
		\includegraphics[interpolate,width=1\linewidth]{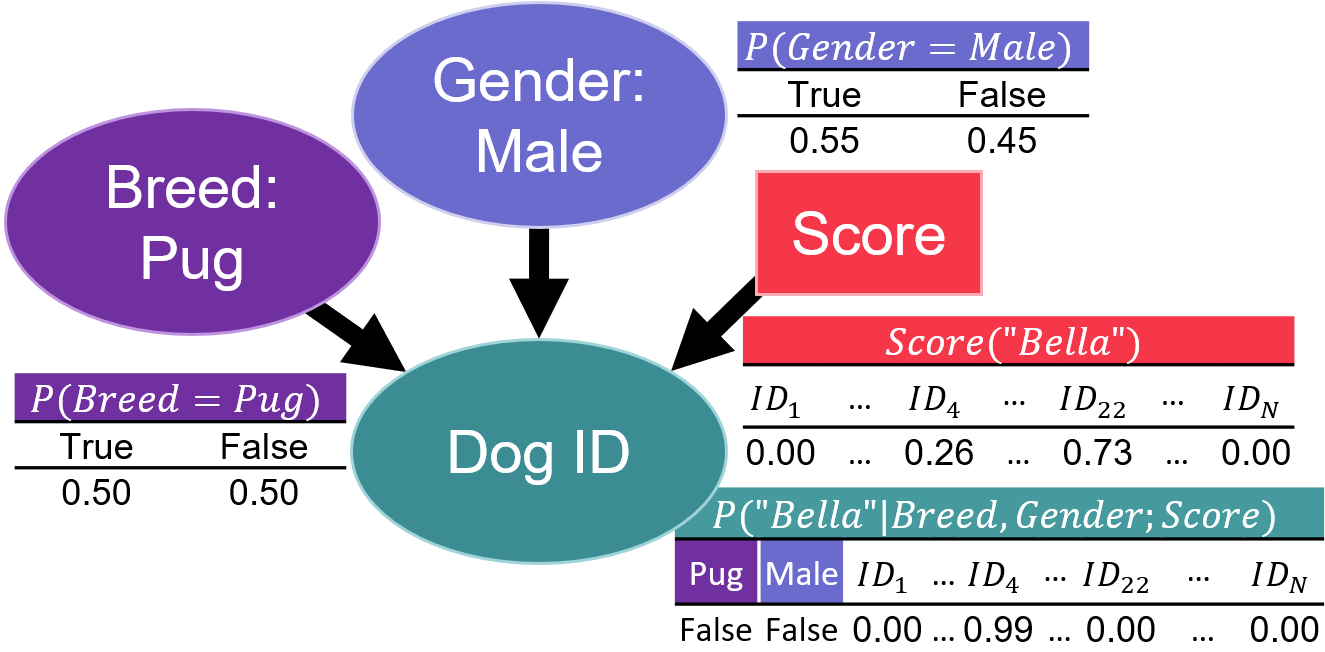}
	\end{center}
	\caption{Example of a decision network for improved dog identification using the two ``soft'' biometrics: breed and gender.}
	\label{figure:bn}
\end{figure}

We can further improve the process of identification by using a ``fusion'' on a classifier decision.  Figure \ref{figure:bn} illustrates a simple decision network containing nodes representing the ``soft'' biometrics (breed and gender) and ``hard'' biometrics (identity).
There is also a  decision node called $Score$. This corresponds to scores that range between 0 and 1 and represent the classifier's confidence in the match. The scores are produced by each classifier; since there are $N$  classes, the vector of scores has  $N$ values.

The conditional probability tables shown in this Figure are calculated based on the distribution of dog breeds and genders in the Flickr-dog dataset.  For example, given 42 dog identities, the chance of the predicted identity being correct is $1/42$.  By knowing the dog is a male pug, the chance of correct prediction becomes $1/14$ where 14 is the total number of dogs that are male pugs.

 This decision network illustrates how the decision is made using the ``soft'' biometrics that is assumed to be known for a probe dog image, and the $Score$ produced for this probe image by the classifier.  Given the classifier's output, $Score$, the score  is modified based on the known information (gender and breed), and a likelihood of the  identity prediction is computed as follows:
	\begin{equation} \label{eq:coeff}
P(id|g,b;s)=\frac{Score(id)\times Ind_{G}(id;g)\times Ind_{B}(id;b)}{Z(g,b;s)} 
	\end{equation}
	
In this Equation, for each probe entry $id\in Val(DogID)$,  $g\in Val(Gender)$, and $b\in Val(Breed)$, where $Val()$ denotes the set of all possible assignments, $Score(id)$ is the matching score, and $Z(g,b;s)$ is the normalization factor. Also, $Ind_{G}(id;g)$ and $Ind_{B}(id;b)$ are the indicators defined as:
 \begin{equation*} \label{eq:indicator}
 \begin{split}
 Ind_{G}(id;g)&=\left\{\begin{array}{ll}1 & \text{\textit{if identity matches the gender}}\\ 0 & \text{\textit{otherwise}}\\ \end{array} 
 \right. \\
 Ind_{B}(id;b)&=\left\{\begin{array}{ll}1 & \text{\textit{if identity matches the breed}}\\ 0 & \text{\textit{otherwise}}\\ \end{array} 
 \right.
 \end{split}
 \end{equation*}

For example, let us assume that given an image of ``Eve'' (husky, female),  the classifier predicts identity as ``Rave'' (husky, male) with high score and ``Eve'' with low score.  Using the default top rank prediction, the identification would fail due to the discrepancy in gender. Using  Equation \ref{eq:coeff},  the ``Rave'' score is penalized by $Ind_{G}(id;g)=0$ while the ``Eve'' score is not.  This application of penalty allows the ``Eve'' score to become the new top rank prediction.
	
Thus, the ``soft'' biometrics help reduce the likelihood of the classifier prediction if these characteristics do not match. In other words, the output scores are adjusted by applying ``probability weights'' based on the breed and gender information.  The output that matches the correct breed and gender remains unchanged, while outputs with non-matching conditions are greatly penalized.  Using this penalty, the top rank prediction from the network is steered towards predictions with the correct breed and gender.

Table \ref{table:4} shows the improved accuracy obtained by using a decision strategy as defined below.  ``Default'' accuracy is the performance obtained by using the assorted CNNs without using ``soft'' biometrics. ``Assisted'' column reports the performance based on the proposed decision network that adjusts the classifier predictions using  ``soft'' biometrics.  There are three partitions of data used for experimentation: ``Pug'', ``Husky'', and ``Pug \& Husky'' (``P \& H'').  

\begin{table}[!htb]
	\caption{Accuracy of Classification using a Decision Network  on the Flickr-dog dataset.}
	\begin{center}
		\begin{tabular}{c|c||ccccc}
			\parbox[t]{2mm}{\multirow{2}{*}{\rotatebox[origin=c]{90}{Data}}} &  \multirow{2}{*}{Method} &  \multicolumn{2}{c}{Balanced} & \multicolumn{2}{c}{Average} \\
			   &  &  Default  &  Assisted &  Default  &  Assisted \\
			\hline
			\hline
\parbox[t]{2mm}{\multirow{5}{*}{\rotatebox[origin=c]{90}{Husky}}}	&	InceptionV3	&	76.11	&	83.33	&	77.70	&	84.37	\\
	&	MobileNet	&	88.73	&	92.70	&	89.41	&	93.24	\\
	&	VGG-16	&	74.05	&	81.90	&	77.26	&	84.16	\\
	&	Xception	&	83.81	&	88.57	&	84.50	&	88.67	\\
\hdashline											
	&	Mean	&	\textbf{80.68}	&	\textbf{86.63}	&	\textbf{82.22}	&	\textbf{87.61}	\\
\hline											
\parbox[t]{2mm}{\multirow{5}{*}{\rotatebox[origin=c]{90}{Pug}}}	&	InceptionV3	&	56.07	&	66.75	&	62.06	&	71.62	\\
	&	MobileNet	&	70.44	&	78.10	&	74.81	&	82.63	\\
	&	VGG-16	&	63.93	&	69.84	&	65.98	&	72.44	\\
	&	Xception	&	65.75	&	76.98	&	71.58	&	81.40	\\
\hdashline											
	&	Mean	&	\textbf{64.05}	&	\textbf{72.92}	&	\textbf{68.61}	&	\textbf{77.02}	\\
\hline											
\parbox[t]{2mm}{\multirow{5}{*}{\rotatebox[origin=c]{90}{P \& H}}}	&	InceptionV3	&	71.92	&	78.53	&	74.26	&	80.23	\\
	&	MobileNet	&	73.75	&	82.68	&	75.38	&	84.04	\\
	&	VGG-16	&	62.62	&	71.98	&	64.62	&	74.27	\\
	&	Xception	&	76.53	&	83.69	&	78.09	&	84.94	\\
\hdashline											
	&	Mean	&	\textbf{71.21}	&	\textbf{79.22}	&	\textbf{73.09}	&	\textbf{80.87}	\\
		\end{tabular}
	\end{center}
	\label{table:4}
\end{table}

Using MobileNet as a classifier, we achieved an average accuracy of 89.41\% and 74.81\% for huskies and pugs, respectively.  This difference in accuracy between the two breeds of dogs is noted in \cite{moreira2017my} as pugs are more difficult to identify.  This behavior is observed for other CNNs.

Usage of  the proposed decision strategy improves each classifier's performance on average by 5.4, 8.4, and 7.8\% for the Pug, Husky, and P \& H data groups, respectively. In particular, using the Xception model as the classifier in combination with the fusion of ``soft'' biometrics, we can achieve approximately 11.2\% higher accuracy for dog identification when compared to previous works.

\section{Conclusion}

The core idea of the proposed approach is a two-stage process, ``coarse-to-fine approach'', that performs  classification of dog breeds prior to identification of each individual dog. The classification or identification at each stage is performed using CNNs. We proposed using different CNNs to examine the performance of breed classification. Applying breed classification at the first stage selects the top-$k$  breeds given a probe image of a dog, which therefore reduces the search space for the next stage. Overall, the proposed approach of using only CNNs achieves on average 6.2\% higher accuracy than that of the previously reported results on dog identification while attempting to identify both Pug and Husky images.  When evaluating the identification rate for each breed of dog separately, an accuracy of 89.41\% is obtained when using the MobileNet model for identifying only husky images and a 74.81\% accuracy is measured for pug images.

In addition, we further improved this identification rate up to 84.94\% by adjusting the likelihood of the classifier prediction using ``soft'' biometrics. The other potential  improvements include better pre-processing techniques, using fine-tuned and optimized CNN architectures, and using advanced decision-level fusion with additional ``soft'' features.

\section{Acknowledgment}
This work was partially supported by the Natural Sciences and Engineering Research Council of Canada through Discovery Grant ``Biometric Intelligent Interfaces''. X. Tu acknowledges the support of her summer research internship (in the Biometric Technologies Laboratory, University of Calgary, Canada) by MITACS Globalink, Canada, and China Scholarship Council, China.  

{\small
\bibliographystyle{IEEEtran}
\bibliography{submission_example}
}


\end{document}